\pdfoutput=1

\documentclass[letterpaper]{article} % DO NOT CHANGE THIS
\usepackage[preprint]{aaai2027}  % DO NOT CHANGE THIS
\usepackage[hyphens]{url}  % DO NOT CHANGE THIS
\usepackage{graphicx} % DO NOT CHANGE THIS
\usepackage{natbib}  % DO NOT CHANGE THIS AND DO NOT ADD ANY OPTIONS TO IT
\usepackage{caption} % DO NOT CHANGE THIS AND DO NOT ADD ANY OPTIONS TO IT
\usepackage{algorithm}
\usepackage{algorithmic}
\usepackage{amsmath, amssymb}
\usepackage{booktabs}
\usepackage{subcaption}
\usepackage[svgnames]{xcolor}
\definecolor{linknavy}{RGB}{0,45,135}
\usepackage[colorlinks=true,linkcolor=DarkBlue,citecolor=DarkBlue,urlcolor=DarkBlue,
pdftitle={Verifier-Guided Synthetic Augmentation for 3D Human Shape Generation},
pdfauthor={Yuexuan Wu, Yang Xiang, Hamid Laga, Dip Das, Anuj Srivastava, Zhengwu Zhang}]{hyperref}

\usepackage{newfloat}
\usepackage{listings}
\DeclareCaptionStyle{ruled}{labelfont=normalfont,labelsep=colon,strut=off} % DO NOT CHANGE THIS
\floatstyle{ruled}
\newfloat{listing}{tb}{lst}{}
\floatname{listing}{Listing}

\usepackage{booktabs}

\newcommand{\sphere}{\ensuremath{\mathbb{S}}}
\newcommand{\real}{\ensuremath{\mathbb{R}}}

\newcommand{\shapes}{\ensuremath{\mathcal{S}}} % Same symbol as \soThreeSet?

\title{Verifier-Guided Synthetic Augmentation for 3D Human Shape Generation}
\author {
    Yuexuan Wu\textsuperscript{\rm 1},
    Yang Xiang\textsuperscript{\rm 2},
    Hamid Laga\textsuperscript{\rm 3},
    Dip Das\textsuperscript{\rm 1},
    Anuj Srivastava\textsuperscript{\rm 4},
    Zhengwu Zhang\textsuperscript{\rm 2} 
}
\affiliations {
    \textsuperscript{\rm 1}Department of Statistics, University of South Carolina, Columbia, SC, USA\\
    \textsuperscript{\rm 2}Department of Statistics and Operations Research, University of North Carolina at Chapel Hill, Chapel Hill, NC, USA\\
    \textsuperscript{\rm 3}School of Information Technology, Murdoch University, Perth, Australia\\
    \textsuperscript{\rm 4}  Department of Applied Mathematics and Statistics, Johns Hopkins University, Baltimore, MD, USA
}

\begin{document}

\maketitle

\begin{abstract}
Limited training data diversity constrains generative modeling of 3D human bodies: conservative models remain close to observed examples, whereas exploratory models often violate basic body proportions. We introduce a verifier-guided augmentation framework that uses global and mode-local PCA to generate inexpensive candidates, screens them using correspondence-derived skeletal proportions and body-part geometry, and retrains a diffusion model on accepted candidates. Elastic registration provides both the modal structure used by distributed PCA and the dense anatomical correspondence needed for scalable screening without per-candidate body-model fitting. A blinded human study supports the verifier as a conservative gatekeeper, favoring verifier-accepted over rejected outputs. We evaluate full-pool verifier acceptance separately from the coverage and departure of accepted samples and combine them through EAUC. On 4,498 registered DFAUST surfaces, distributed-PCA augmentation achieves 86.32\% acceptance, the highest CP-AUC (0.871), and the highest EAUC (0.752), improving EAUC by 32\% over real-only and self-augmented diffusion. These results show that mode-local, verifier-guided proposals broaden diffusion generation while maintaining high agreement with calibrated body measurements.
\end{abstract}


\section{Introduction}

Three-dimensional human surfaces support applications in vision, graphics, virtual reality, digital health, ergonomics, and apparel design. Early scan-based statistical models learned low-dimensional spaces from registered human bodies~\cite{allen2003space}, while articulated models such as SCAPE and SMPL represent body shape and pose compactly~\cite{anguelov2005scape,loper2015smpl}. Neural alternatives include mesh VAEs and geometry-aware latent models~\cite{tan2018meshvae,bouritsas2019neural,cosmo2020limp}, parametric and implicit human models~\cite{xu2020ghum,alldieck2021imghum}, and recent generators for articulated, textured, or clothed 3D humans~\cite{chen2023primdiffusion,yoshiyasu2024diffsurf,seth2025pocoloco,ji2025jade,yoshiyasu2025meshmamba,tang2026geometry}. Yet data remain a bottleneck. DFAUST, for example, contains many motion frames but few identities~\cite{bogo2017dfaust}, so a generator may fit the recorded sample without spanning the broader population of human body shapes. Synthetic augmentation is attractive, but near-reference samples may add little diversity, while exploratory samples may violate selected body statistics.

Registered PCA models generate inexpensive, far-reaching proposals, whereas real-only diffusion yields high verifier acceptance but remains close to the observed PCA-score distribution. We therefore use verifier screening to turn exploratory PCA proposals into augmentation data that broadens diffusion while controlling which candidates enter training.

Elastic shape analysis enables this design. Elastic metrics compare surfaces while accounting for rigid motion and reparameterization~\cite{jermyn2012srnf,jermyn2017elastic,hartman2023elastic,pierson2022riemannian}, and common-template registration establishes dense correspondence. We use this geometry for registration and mode discovery, then fit coordinate-PCA models globally and within modal clusters~\cite{deng2022characterizing}. A global model can mix variation across modes; local models restrict sampling to modal neighborhoods. Elastic geometry thus supplies sample structure and correspondence, while PCA supplies low-dimensional proposals. Correspondence also lets fixed anatomical neighborhoods and part labels be reused for screening. Building on prior elastic surface generation and human-body analysis~\cite{jermyn2017elastic,hartman2023baresa}, we use these capabilities for verifier-guided augmentation.

\begin{figure}[t]
    \centering
    \includegraphics[width=\linewidth]{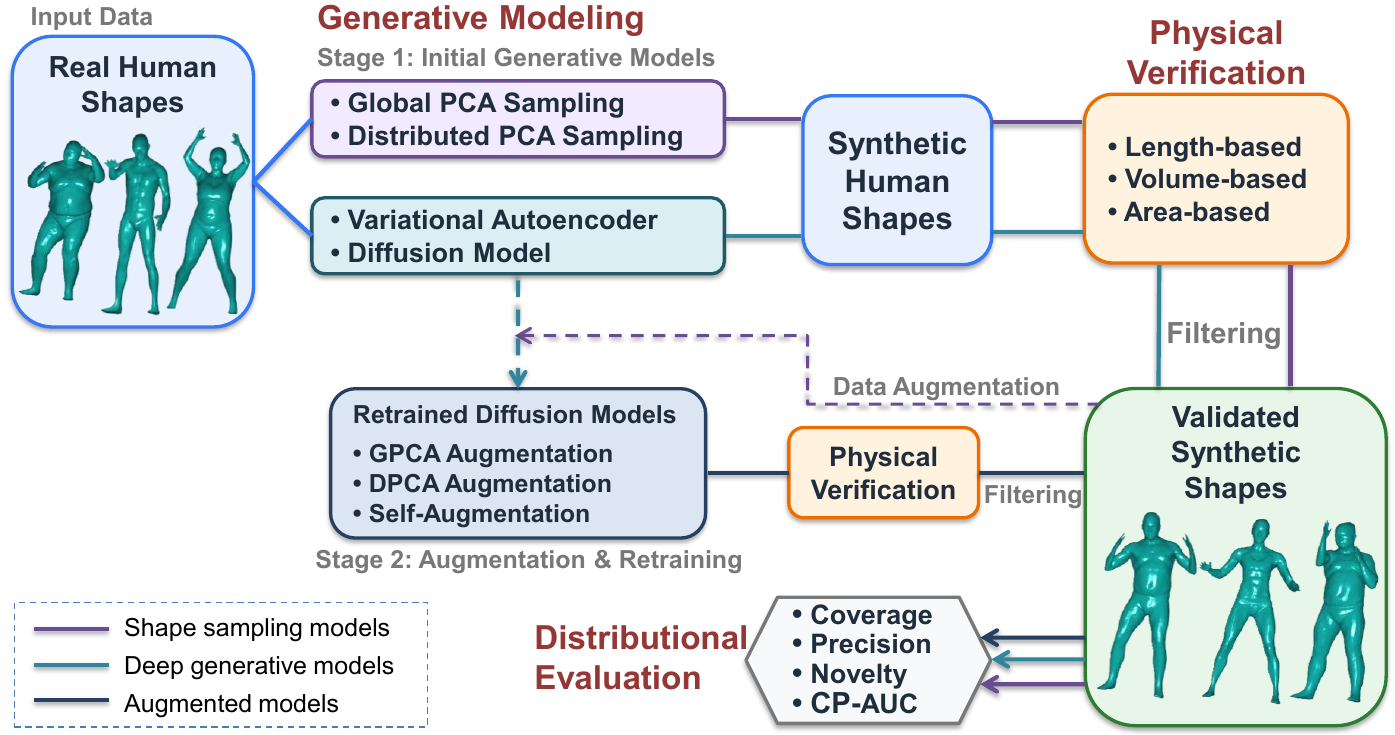}
    \caption{Two-round verifier-guided augmentation pipeline. GPCA, DPCA, VAE, and real-only diffusion models are first fitted to the real surfaces and evaluated after reference-calibrated screening. Verifier-accepted GPCA, DPCA, and real-only diffusion samples form equal-size augmentation pools; each is combined with the real surfaces to train a new diffusion model from scratch. The VAE remains a first-round comparator.}
    \label{fig:pipeline}
\end{figure}

Our framework (Fig.~\ref{fig:pipeline}) separates two rounds of model fitting. First, we fit GPCA, DPCA, a VAE, and a real-only diffusion; GPCA and DPCA provide exploratory pools, while diffusion provides self-augmentation control. A reference-calibrated verifier screens skeletal proportions, body-part volumes, and arm cross-sectional areas. Accepted candidates are then combined with the real surfaces to train diffusion models from scratch. Dense correspondence makes screening inexpensive, and equal accepted-pool sizes enable a controlled comparison of augmentation sources. This filter-before-reuse design is motivated by the risks of self-consuming training~\cite{alemohammad2024self}, sample-level audits of synthetic outputs~\cite{alaa2022faithful}, and analyses showing that curation or verification can make synthetic data more useful for retraining~\cite{ferbach2024selfconsuming,feng2025beyond}. We instantiate this principle for registered human surfaces using reference-calibrated body statistics, building on the broader use of anthropometric and physical constraints in human-shape and pose estimation~\cite{choutas2022shapy,akhter2015pose,tripathi20233d}. We also complement it with a blinded human annotation to test whether verifier enriches for human-perceived validity.

Verifier acceptance characterizes individual shapes, while complementary distributional analysis measures how a generated collection covers and extends the observed population. Set-based evaluation of generated point clouds pairs matching-based fidelity with coverage~\cite{achlioptas2018pointclouds}, while broader generative-model metrics separate fidelity and diversity using precision and recall or density and coverage~\cite{sajjadi2018assessing,kynkaanniemi2019improved,naeem2020reliable}. Motivated by this distinction, we report full-pool acceptance and define radius-swept coverage, precision, and novelty on equal-size accepted sets in a common PCA space. CP-AUC summarizes the coverage--precision curve, and EAUC combines CP-AUC with acceptance. Diffusion trained with accepted DPCA samples attains 86.32\% acceptance, the highest CP-AUC (0.871), and the highest EAUC (0.752), versus EAUC 0.568 for the same-size self-augmentation control. These results identify mode-local, verifier-guided augmentation as the strongest strategy in our comparison.

Our contributions are threefold:
\begin{itemize}
    \item We integrate elastic registration and mode discovery with global and mode-local PCA proposals, correspondence-enabled screening, and diffusion augmentation into a single, controlled proposal--verification--augmentation pipeline.
    \item We construct a reference-calibrated shape verifier from correspondence-derived skeletal proportions and body-part geometry, enabling inexpensive screening of large candidate pools, and assess its decisions through a blinded rating study.
    \item We separate acceptance from nearest-neighbor coverage and novelty and include a same-size self-augmentation control; DPCA-augmented diffusion attains the highest CP-AUC and EAUC.
\end{itemize}

\section{Datasets and Data Representations}

\subsection{DFAUST Dataset}
We use 4,498 DFAUST surface scans from 17 punching and jumping-jack trajectories by 10 subjects: seven perform both motions and three perform punching only~\cite{bogo2017dfaust}. We retain every frame and treat it as an independent surface; differing trajectory durations make subject and motion counts unequal.

\subsection{Elastic Shape Representations and Analysis}

Each closed genus-zero mesh is mapped to \(\sphere^2\) using spherical conformal and Tutte mappings~\cite{jermyn2017elastic}, yielding \(f_i:\sphere^2\rightarrow\real^3\), \(i=1,\ldots,N\), evaluated at \(P=256^2\) common grid locations. This supplies a common domain; anatomical correspondence is subsequently estimated through elastic reparameterization.

\paragraph{Elastic Representation and Registration.}
Let $\mathcal{F}$ denote the space of parameterized surfaces
$f:\sphere^2\rightarrow\real^3$, and let $\Gamma$ be the set of
orientation-preserving diffeomorphisms of $\sphere^2$. We write
\(\shapes=\mathcal F/(SO(3)\times\Gamma)\) for the corresponding shape space and
\([f]\in\shapes\) for the rotation--reparameterization equivalence class of \(f\).
For \(\omega=(u,v)\in\sphere^2\), let
\(
n_f(\omega) = \partial_u f(\omega) \times \partial_v f(\omega)
\)
denote the unnormalized surface normal. Following \citet{jermyn2012srnf}, the square-root normal field (SRNF) representation of surface $f$ is defined as
\(
    q_f(\omega) = n_f(\omega)/\sqrt{\lVert n_f(\omega) \rVert_2}\,\,.
\)
For any field \(q:\sphere^2\to\real^3\) and reparameterization $\gamma\in\Gamma$, define $q\star\gamma=(q\circ\gamma)\sqrt{J_\gamma}$, where $J_\gamma$ is the area Jacobian of $\gamma$. The (extrinsic) elastic shape distance between two surfaces is then
\begin{equation}
    d_{\shapes}([f_1],[f_2]) =
    \inf_{O \in SO(3), \gamma\in\Gamma}
    \left\| q_{f_1} - O\left( q_{f_2}\star\gamma \right)
    \right\|_{L^2(\sphere^2,\real^3)}.
    \label{eq:dist_shape}
\end{equation}
The joint minimization over rotation $O$ and reparameterization $\gamma$ aligns the surfaces and establishes dense correspondence. Thus, registration is estimated jointly with elastic shape comparison rather than as a separate preprocessing step.

\paragraph{Population Template and Global PCA.}
We target the Fr\'echet-mean objective under the elastic distance in Eq.~\eqref{eq:dist_shape}:
\(
[\bar f]\in \arg\min_{[f]\in\shapes} \sum_{i=1}^{N}
    d_{\shapes}^{2}([f],[f_i]).
\)

We choose a coordinate representative \(\bar f\) of the minimizing class and use an efficient alternating approximation: surfaces are registered to the current template, after which the template is updated by averaging the registered coordinates. At convergence, let \((O_i^*,\gamma_i^*)\) denote the approximate minimizers of Eq.~\eqref{eq:dist_shape} relative to \(\bar f\), and define
\(
    f_i^{*}(\omega) = O_i^{*} f_i\!\left(\gamma_i^{*}(\omega)\right)
\)
to be surface \(f_i\) registered to \(\bar f\).

Let \(p=3P\). We represent each aligned surface by the ambient coordinate deformation
\(\mathbf v_i=\operatorname{vec}(f_i^{*}-\bar f)\in\real^p\), where \(\operatorname{vec}\) stacks the three coordinates at the \(P\) grid locations and \(\operatorname{unvec}\) is its inverse. Applying PCA to \(\{\mathbf v_i\}_{i=1}^{N}\) gives an orthonormal basis \(\Phi_G\in\real^{p\times d_G}\) and global score vectors
\(
    \mathbf z_i=\Phi_G^\top\mathbf v_i\in\real^{d_G}.
\)
We set \(d_G=25\), retaining more than \(99\%\) of the registered-coordinate variance, and reconstruct a surface from any global score \(\mathbf z\in\real^{d_G}\) by
\begin{equation}
    \widehat f(\mathbf z) = \bar f + \operatorname{unvec}(\Phi_G\mathbf z).
    \label{eq:recon}
\end{equation}
We call this representation global PCA (GPCA) and use its scores for neural modeling and distributional evaluation.

Elastic distances, rather than Riemannian logarithm maps, supply alignment, dense correspondence, and clustering; PCA of ambient coordinate deformations supplies the finite-dimensional model for sampling and neural learning. Because Eq.~\eqref{eq:recon} preserves registered vertex ordering, fixed anatomical neighborhoods can be evaluated without fitting a body model to every candidate.

\paragraph{Mode-Based Clustering and Distributed PCA.}

\begin{figure} 
    \centering
    \includegraphics[width=\linewidth]{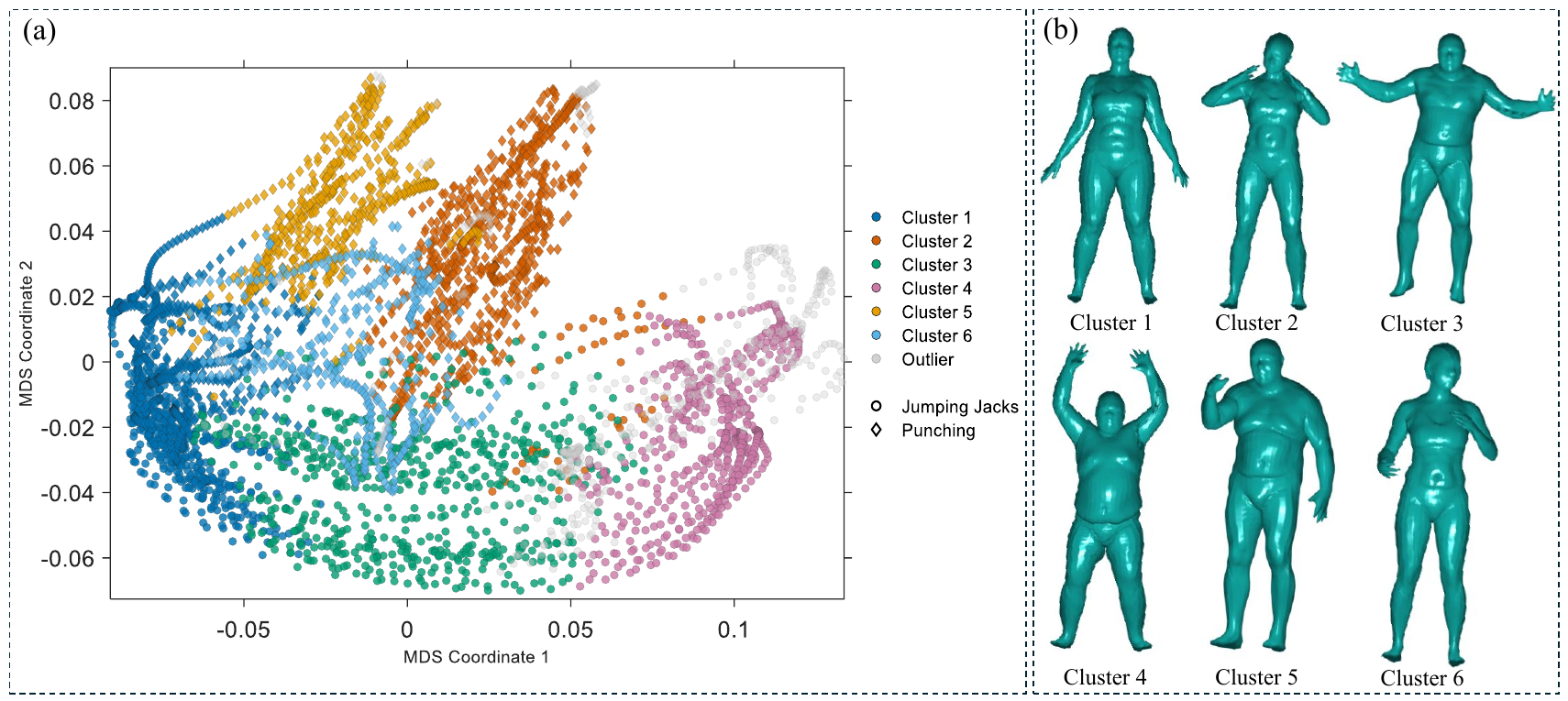}
    \caption{Mode-based structure of the registered DFAUST population. (a) 2D MDS of pairwise elastic shape distances, colored by the six clusters; gray denotes observations classified as outliers. Circles denote jumping jacks; diamonds denote punching. (b) Modal shape for each cluster.}
    \label{fig:distributed_pca}
\end{figure}

GPCA imposes one linear approximation around a global template; for a multimodal sample, its directions can mix within-mode variability with differences between modes. We therefore apply the $k$-mode kernel-mixture method~\cite{deng2022characterizing} to the pairwise elastic distances and obtain \(C=6\) modal clusters, treating observations outside their neighborhoods as outliers. Fig.~\ref{fig:distributed_pca} shows a two-dimensional MDS visualization and the corresponding mode surfaces; clustering itself uses the full distance matrix. The clusters are visibly organized by motion type: as read from Fig.~\ref{fig:distributed_pca}(a), clusters~3 and~4 consist of jumping jack frames, cluster~5 and most of cluster~2 of punching frames, and cluster~1 and cluster~6 contain both motions, so the mode structure reflects pose and motion in addition to body morphology.

For each \(c\in\{1,\ldots,C\}\), let \(\mathcal I_c\subseteq\{1,\ldots,N\}\) be the indices assigned to cluster \(c\). We repeat alternating template estimation and registration within \(\mathcal I_c\) and fit a coordinate-PCA model to the resulting local deformations. Let \(f_i^{*(c)}\) denote surface \(i\) after this cluster-local registration, let \(\bar f_c\) be the local template, and let \(\Phi_c\in\real^{p\times d_c}\) be the local basis, where \(d_c\) is the retained dimension for cluster \(c\). The corresponding local score is
\(\mathbf a_i^{(c)}=\Phi_c^\top\operatorname{vec}(f_i^{*(c)}-\bar f_c)\in\real^{d_c}\).
Observations outside the modal neighborhoods are excluded from local fitting. We call the \(C\)-model collection distributed PCA (DPCA). It is related to local PCA and probabilistic-PCA mixtures~\cite{kambhatla1997localpca,tipping1999mppca}, but its elastic-distance clusters serve as balanced proposal sources rather than a fitted mixture. Generated surfaces are subsequently encoded in the \(d_G=25\)-dimensional global space to provide a common neural input and evaluation representation.

\section{First-Round Generative Models}
Before augmentation, we fit GPCA and DPCA proposal models and VAE and diffusion baselines to the real surfaces; the neural models share an 80/20 score split. Verifier-accepted GPCA, DPCA, and real-only diffusion samples form the second-round pools, while the VAE remains a first-round comparator.

\paragraph{Global PCA Proposal Model.}
Let \(\widehat{\boldsymbol\mu}_G,\widehat{\boldsymbol\sigma}_G\in\real^{d_G}\) be the component-wise empirical mean and standard deviation of the global scores. We draw
\(\widetilde{\mathbf z}=\widehat{\boldsymbol\mu}_G+\widehat{\boldsymbol\sigma}_G\odot\boldsymbol\epsilon_G\), where
\(\boldsymbol\epsilon_G\sim\mathcal N(\mathbf 0,I_{d_G})\), \(I_d\) denotes the \(d\times d\) identity matrix, and \(\odot\) denotes component-wise multiplication. We reconstruct with Eq.~\eqref{eq:recon}, producing 100,000 GPCA candidates.

\paragraph{Distributed PCA Proposal Models.}
For cluster \(c\), let \(\widehat{\boldsymbol\mu}_c,\widehat{\boldsymbol\sigma}_c\in\real^{d_c}\) denote the component-wise mean and standard deviation of \(\{\mathbf a_i^{(c)}:i\in\mathcal I_c\}\). We draw
\(\widetilde{\mathbf a}^{(c)}=\widehat{\boldsymbol\mu}_c+\widehat{\boldsymbol\sigma}_c\odot\boldsymbol\epsilon^{(c)}\), where \(\boldsymbol\epsilon^{(c)}\sim\mathcal N(\mathbf 0,I_{d_c})\), and reconstruct
\(\widetilde f^{(c)}=\bar f_c+\operatorname{unvec}(\Phi_c\widetilde{\mathbf a}^{(c)})\).
Sampling 7,500 candidates from each of the \(C=6\) modes yields 45,000 balanced DPCA proposals; equal allocation ensures that every detected mode can contribute to augmentation. Accepted surfaces are encoded in the global \(d_G\)-dimensional space used by all neural models and evaluation metrics.

\paragraph{Variational Autoencoder Baseline.}
The VAE~\cite{kingma2014autoencoding} provides a nonlinear latent-variable baseline for the first-round comparison; it is not used as an augmentation source or recipient. It has four fully connected layers of width 64 in both encoder and decoder and an eight-dimensional Gaussian latent space. The KL weight increases linearly from \(10^{-3}\) to \(0.1\) over 300 epochs. We train for 400 epochs with Adam~\cite{kingma2015adam} (batch size 128 and learning rate \(10^{-3}\)). For generation, we draw latent vectors from \(\mathcal N(\mathbf 0,I_8)\), decode them to global PCA scores, reconstruct surfaces using Eq.~\eqref{eq:recon}, and retain 40,000 samples.

\paragraph{Diffusion Model Trained on Real Data.}
We train a denoising diffusion probabilistic model~\cite{ho2020denoising} on the real \(d_G=25\)-dimensional global PCA scores. Training observations are shuffled within each epoch, whereas validation observations are not. The forward process uses \(T=1000\) steps and a linear variance schedule from \(\beta_1=10^{-4}\) to \(\beta_T=0.02\). Let \(\mathbf z_0\in\real^{d_G}\) denote a clean global PCA score; Eq.~\eqref{eq:recon} maps this same type of score to a registered surface. With \(\alpha_t=1-\beta_t\) and \(\bar\alpha_t=\prod_{s=1}^{t}\alpha_s\), its noisy state at diffusion step \(t\) is
\(
\mathbf z_t=\sqrt{\bar\alpha_t}\mathbf z_0+
\sqrt{1-\bar\alpha_t}\boldsymbol\epsilon \text{, with }
\boldsymbol\epsilon\sim\mathcal N(\mathbf 0,I_{d_G}).
\)

The denoiser predicts the added Gaussian noise using the standard DDPM objective~\cite{ho2020denoising}. A U-Net-style network~\cite{ronneberger2015unet} processes each score vector reshaped to \(5\times5\), using a 32-dimensional time embedding, widths 128 and 256, and up to 512 bottleneck channels. We train for 1,000 epochs with Adam~\cite{kingma2015adam} (batch size 128, learning rate \(10^{-3}\)), retain the lowest-validation-loss checkpoint, and reconstruct reverse-process samples through Eq.~\eqref{eq:recon}. The model generates 40,000 samples for evaluation and self-augmentation.

\section{Reference-Calibrated Shape Verifier}
The verifier extracts correspondence-based skeletal proportions, part-volume ratios, and arm-area proxies from a registered surface and returns a binary screening decision. Dense correspondence makes these features inexpensive enough to screen large augmentation pools.

\subsection{Correspondence-Based Verifier Features}

\paragraph{Skeleton-Based Proportion Features.}
Rather than fit a complex SMPL model to every candidate shape, we use it only during a one-time joint-supervision stage. We transform registered surfaces to the SMPL template space~\cite{loper2015smpl}, fit pose, orientation, and translation for a 120-surface real joint-supervision subset, and map the resulting 24 joints back to registered-surface coordinates.

This subset is used to supervise correspondence-based joint predictors. Let \(\mathcal T\subseteq\{1,\ldots,N\}\) be the index set of the 120 joint-supervision surfaces. For joint \(k\in\{1,\ldots,24\}\), let \(S_k\subseteq\{1,\ldots,P\}\) contain the \(h=40\) grid indices nearest to that joint on the reference template. Dense correspondence gives the indices in \(S_k\) a consistent meaning on every surface. For an arbitrary registered surface \(f\), let \(\mathbf u_k(f)\in\real^{3h}\) vectorize the coordinates indexed by \(S_k\). On joint-supervision surface \(i\), write \(\mathbf u_{i,k}=\mathbf u_k(f_i^*)\), and let \(\mathbf j_{i,k}\in\real^3\) be the fitted joint coordinate. We estimate
\begin{equation*}
\begin{aligned}
(\widehat W_k,\widehat{\mathbf b}_k)
&=
\arg\min_{W_k,\mathbf b_k}
\sum_{i\in\mathcal T}
\left\|
\mathbf j_{i,k}
-
W_k^\top\mathbf u_{i,k}
-
\mathbf b_k
\right\|_2^2 \\
&\qquad
+
\lambda\left\|W_k\right\|_F^2 ,
\end{aligned}
\end{equation*}
where \(W_k\in\real^{3h\times3}\), \(\mathbf b_k\in\real^3\), and \(\lambda\ge0\) is the ridge parameter. For any surface $f$, we predict
\(
\widehat{\mathbf j}_k(f)=\widehat W_k^\top\mathbf u_k(f)+\widehat{\mathbf b}_k
\). Once calibrated, the 24 predictors require only linear evaluation of fixed neighborhoods. Their predicted joints provide left--right arm and leg symmetry ratios, arm-to-leg ratios, and forearm-to-upper-arm ratios without repeated SMPL optimization.

\paragraph{Part-Volume Ratio Features.}
For each surface \(f\) reconstructed from its \(d_G=25\)-dimensional global PCA score through Eq.~\eqref{eq:recon}, we align it to a fixed, segmented SMPL template. We scale the aligned surface to the template height and rescale only its second coordinate to match the corresponding template extent. Anatomical labels are transferred without per-candidate SMPL fitting: each reconstructed point receives the majority vertex label of the closest template triangle under Euclidean point-to-triangle distance.

For anatomical part \(k\), let \(\mathcal P_k(f)\subset\real^3\) contain its assigned points. Because they need not form watertight submeshes, we use the convex-hull volume as a deterministic proxy:
$
\widehat V_k(f)=
\operatorname{Vol}\!\left(\operatorname{ConvHull}\!\left(\mathcal P_k(f)\right)\right).
$
Composite proxies sum segment volumes, e.g.,
\(\widehat V_{\mathrm{left\,arm}}=\widehat V_{\mathrm{left\,upper\,arm}}+\widehat V_{\mathrm{left\,forearm}}\), with analogous definitions for the legs and right limbs. Their left/right arm and leg, arm/leg, and head/arm ratios characterize relative proportions rather than body size.

\paragraph{Arm Cross-Sectional-Area Features.}
Let \(\mathcal A\) index the left and right upper arms and forearms. For \(a\in\mathcal A\), let \(\widehat L_a(f)>0\) be the distance between the predicted endpoint joints of arm segment \(a\) on surface \(f\). To target abnormally thin or enlarged arms, we define
\(\widehat A_a^{\mathrm{eff}}(f)=\widehat V_a(f)/\widehat L_a(f)\).
This length-normalized volume approximates the average cross-sectional area.

\paragraph{Feature Inventory.}
In total, the verifier evaluates three groups of scalar features, each containing four features: the skeletal length ratios (left/right arm, left/right leg, arm/leg, and forearm/upper-arm), the part-volume ratios (left/right arm, left/right leg, arm/leg, and head/arm), and the arm cross-sectional-area proxies (left and right upper arms and forearms).

\subsection{Reference Calibration and Screening Rule}
The 120-surface subset fits only the joint predictors; thresholds use the real training split. For each feature \(g_\ell\), let \(\mu_\ell^{\mathrm{ref}}\) and \(\sigma_\ell^{\mathrm{ref}}>0\) be its training mean and standard deviation, define
\(\xi_\ell(f)=(g_\ell(f)-\mu_\ell^{\mathrm{ref}})/\sigma_\ell^{\mathrm{ref}}\), and accept when \(\max_\ell|\xi_\ell(f)|\le3\).
Length, volume, and arm-area checks are applied sequentially, and a shape is accepted when it agrees with all calibrated skeletal and body-part criteria. This provides a consistent, reproducible screening rule for large synthetic pools.

\section{Verifier-Guided Diffusion Augmentation}
Only verifier-accepted candidates enter second-round training.

\paragraph{Why Diffusion Is the Augmentation Recipient.}
Every accepted surface shares the global PCA representation, so diffusion provides a fixed recipient while the synthetic source varies. Real-only diffusion has high acceptance but limited departure, whereas PCA proposals explore more broadly with more rejections; the second round tests whether screened proposals can broaden diffusion. Accordingly, diffusion is the sole second-round recipient, while the VAE serves as a first-round neural comparator.

\paragraph{Augmentation-Pool Construction.}
For each of the three augmentation sources (GPCA, DPCA, and the real-only diffusion model), we uniformly select 12,000 verifier-accepted surfaces without replacement and combine them with all 4,498 real surfaces. These three datasets define Diffusion-GPCA, Diffusion-DPCA, and Diffusion-SelfAug, respectively.
The DPCA candidate pool is generated with equal allocation across clusters before screening; the 12,000 retained DPCA surfaces are then drawn uniformly from the pooled accepted candidates, so their post-screening cluster counts need not be equal.

\paragraph{Controlled Diffusion Retraining.}
Each 16,498-example collection is divided into training and validation sets using the same 80/20 procedure. These models constitute a second round of fitting: every augmented diffusion model is trained from random initialization rather than fine-tuned from the real-only checkpoint. We use the same architecture, diffusion schedule, loss, optimizer, learning rate, batch size, epoch budget, and checkpoint-selection rule for all three variants. The three augmented variants therefore share data volume and optimization settings, enabling a controlled comparison of augmentation sources; the real-only diffusion model provides the unaugmented reference. Each augmented diffusion variant generates 40,000 evaluation samples.

\section{Evaluation Protocol}
We evaluate two complementary properties: verifier acceptance over full output pools and distributional behavior for equal-size accepted sets in the common global PCA space. Acceptance can be high for a narrow distribution, while broad coverage or departure does not indicate rejection rates.

\subsection{Verifier Acceptance Rate}
For full output set \(\mathcal G\) and accepted subset \(\mathcal G_{\mathrm{acc}}\), define
\(\rho_{\mathrm{acc}}=|\mathcal G_{\mathrm{acc}}|/|\mathcal G|\).
This rate quantifies agreement with the calibrated body criteria across the complete output pool.

\subsection{Distributional Evaluation of Accepted Shapes}
\paragraph{Common Representation and Nearest Neighbors.}
To compare all methods in a common representation, we encode every real and generated surface in the global \(d_G\)-dimensional PCA space. Let
\(X=\{\mathbf x_i\}_{i=1}^{N_X}\subset\real^{d_G}\), with \(N_X=4{,}498\), contain all real shape scores. We use them as the reference population to measure agreement with the complete observed sample; this is a distributional comparison rather than a held-out generalization test. Let
\(Y=\{\mathbf y_j\}_{j=1}^{N_Y}\subset\real^{d_G}\), with \(N_Y=10{,}000\), be a uniform draw without replacement from a method's accepted scores. We define
\(
\delta_{X\to Y}(\mathbf x_i)
=\min_{\mathbf y\in Y}\|\mathbf x_i-\mathbf y\|_2
\) and 
\(
\delta_{Y\to X}(\mathbf y_j)
=\min_{\mathbf x\in X}\|\mathbf y_j-\mathbf x\|_2.
\)
\paragraph{Coverage, Precision, and Novelty.}
For radius \(r\ge0\),
\begin{align}
\operatorname{Cov}(r)
&=\frac{1}{N_X}\sum_{i=1}^{N_X}
\mathbf 1\{\delta_{X\rightarrow Y}(\mathbf x_i)\le r\},
\label{eq:Coverage}\\
\operatorname{Prec}(r)
&=\frac{1}{N_Y}\sum_{j=1}^{N_Y}
\mathbf 1\{\delta_{Y\rightarrow X}(\mathbf y_j)\le r\},
\label{eq:Precision}\\
\operatorname{Nov}(r)&=1-\operatorname{Prec}(r).
\label{eq:Novelty}
\end{align}
Coverage and precision are the empirical CDFs of the two directional distances; persistent novelty indicates greater accepted-sample departure from the reference.

\paragraph{Coverage--Precision Summary.}
We plot the threshold sweep \(r\mapsto(\operatorname{Prec}(r),\operatorname{Cov}(r))\) and report
\begin{equation*}
\operatorname{CP\text{-}AUC}
=\int_{[0,\infty)}\operatorname{Cov}(r)\,d\operatorname{Prec}(r).
% \label{eq:cp_auc}
\end{equation*}
For independent uniform draws \(D_{X\rightarrow Y}\) and \(D_{Y\rightarrow X}\) from the two empirical distance collections, CP-AUC equals
\(\Pr[D_{X\rightarrow Y}\le D_{Y\rightarrow X}]\), up to ties. As a directional rank statistic, CP-AUC jointly reflects coverage and departure, so we interpret it with the distance curves and quantile radii.

\paragraph{Quantile Radii and Cardinality Context.}
For \(\tau\in(0,1)\), define
\(r_\tau^{\mathrm{cov}}=\inf\{r\ge0:\operatorname{Cov}(r)\ge\tau\}\) and
\(r_\tau^{\mathrm{nov}}=\sup\{r\ge0:\operatorname{Nov}(r)\ge\tau\}\).
Percentage subscripts \(r_{50}\), \(r_{70}\), and \(r_{90}\) denote \(\tau=0.50\), \(0.70\), and \(0.90\). Random 80/20 and 20/80 target/pseudo-generated real partitions in Fig.~\ref{fig:distributional_evaluation_with_ref} provide finite-sample context. Table~\ref{tab:real_split_ref} gives the full sweep (0.122 at 95/5 to 0.872 at 5/95), showing sensitivity to relative set sizes.

\subsection{Acceptance-Adjusted Performance}
To combine full-pool acceptance with the fixed-size accepted set comparison, we report the effective AUC (EAUC):
\begin{equation}
\operatorname{EAUC}=\rho_{\mathrm{acc}}\times\operatorname{CP\text{-}AUC}.
\label{eq:eauc}
\end{equation}
EAUC combines full-pool acceptance with accepted-set CP-AUC, and both components are reported alongside it.

\section{Experimental Results}

\subsection{Verifier Acceptance Results}

We apply the reference-calibrated verifier to each real and generated population. Table~\ref{tab:filtering_results} summarizes the final verifier acceptance counts and rates. Table~\ref{tab:filtering_results_full} in the Appendix reports detailed rejection counts at three screening stages.

\begin{figure}[t]
    \centering
    \includegraphics[width=.75\linewidth]{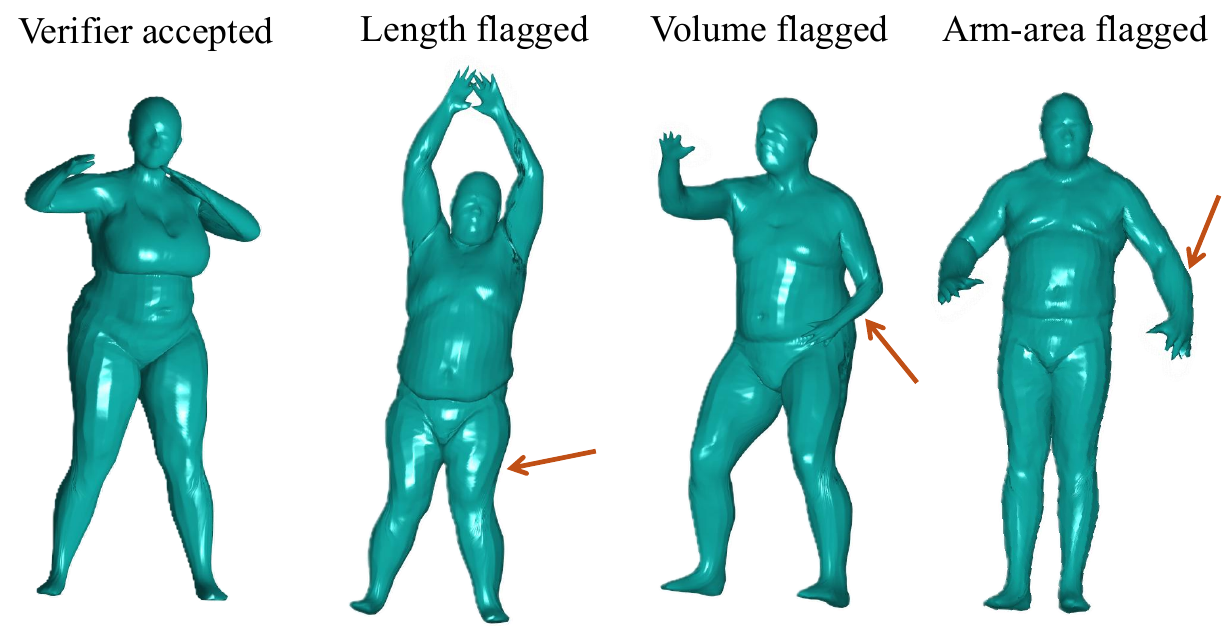}
    \caption{Representative outcomes from the sequential verifier. From left to right: verifier-accepted; length-flagged; volume-flagged; and arm-area-flagged shapes. Arrows indicate the observed abnormal regions.}
    \label{fig:example_filtration}
\end{figure}

Fig.~\ref{fig:example_filtration} illustrates representative Diffusion-DPCA outcomes from the sequential verifier. Within each accepted or newly flagged group, we select the shape whose \(\max_\ell|\xi_\ell(f)|\) is closest to the group median. The three rejection stages identify complementary failures: the length-flagged example shows
disproportionately short legs; the volume-flagged example passes the length checks but shows a thin and asymmetric left arm; and the arm-area-rejected example passes both preceding stages but retains an abnormal left-forearm cross-sectional size.

Distributed PCA raises verifier acceptance from 13.82\% for Global PCA to 66.85\%. The VAE reaches 61.71\%, while real-only Diffusion achieves 93.53\%, approaching the 96.87\% real-data reference. 

% \begin{table*}[ht]
% \centering

%     \begin{tabular}{lrrrrr}
%     \toprule
%     \textbf{Filtering metric} & \textbf{Global PCA} & \textbf{Distributed PCA} & \textbf{VAE} & \textbf{Diffusion} & \textbf{Real shapes} \\
%     \midrule
%     Original \# of shapes & 100{,}000 & 45{,}000 & 40{,}000 & 40{,}000 & 4{,}498 \\
%     Flagged by length & 80{,}737 & 11{,}648 & 13{,}304 & 1{,}763 & 58 \\
%     Flagged by volume & 5{,}204 & 2{,}664 & {1,768} & 697 & 80 \\
%     Flagged by area & 238 & 607 & 244  & 129 & 3 \\
%     \midrule
%     \textbf{Verifier accepted} & \textbf{13{,}821} & \textbf{30{,}081} & \textbf{24{,}684} & \textbf{37{,}411} & \textbf{4{,}357} \\
%     \textbf{Acceptance rate} & \textbf{13.82\%} & \textbf{66.85\%} & \textbf{61.71\%} & \textbf{93.53\%} & \textbf{96.87\%} \\
%     \bottomrule

%     \end{tabular}

% \caption{Summary of sequential validity filtering results across real and generated human shapes. Volume-based filtering is applied after removing shapes flagged by length-based metrics; area-based filtering is applied after removing shapes flagged by either length-based or volume-based metrics.}
% \label{tab:filtering_sum}
% \end{table*}

Among augmented models, Diffusion-DPCA achieves 86.32\% acceptance, 15.85 percentage points above Diffusion-GPCA, while retaining the broader-support behavior examined below. Diffusion-SelfAug reaches 94.59\%, and Diffusion-GPCA reaches 70.47\%. Relative to their proposal pools, diffusion retraining raises acceptance from 13.82\% to 70.47\% for GPCA and from 66.85\% to 86.32\% for DPCA.

% \begin{table*}[t] 
% \centering 
% \small
% \begin{tabular}{lrrrrrr} 
% \toprule 
% & \textbf{Initial number} 
% & \multicolumn{3}{c}{\textbf{Flagged by filtering metric}} 
% & \textbf{Verifier} 
% & \textbf{Acceptance} \\
% \cmidrule(lr){3-5} \textbf{Source} & \textbf{of shapes} & \textbf{Length} & \textbf{Volume} & \textbf{Area} & \textbf{accepted} & \textbf{rate (\%)} \\
% \midrule 

% \multicolumn{7}{l}{\textit{Base generators}} \\ 
% Global PCA & 100{,}000 & 80{,}737 & 5{,}204 & 238 & 13{,}821 & 13.82 \\ Distributed PCA & 45{,}000 & 11{,}648 & 2{,}664 & 607 & 30{,}081 & 66.85 \\ 
% VAE & 40{,}000 & 13{,}304 & 1{,}768 & 244 & 24{,}684 & 61.71 \\
% Diffusion & 40{,}000 & 1{,}763 & 697 & 129 & 37{,}411 & 93.53 \\
% \midrule 

% \multicolumn{7}{l}{\textit{Augmented diffusion generators}} \\
% Diffusion-GPCA & 40{,}000 & 9{,}250 & 2{,}285 & 279 & 28{,}186 & 70.47 \\
% Diffusion-DPCA & 40{,}000 & 3{,}875 & 1{,}243 & 355 & 34{,}527 & 86.32 \\ Diffusion-SelfAug & 40{,}000 & 1{,}044 & 1{,}041 & 78 & 37{,}837 & 94.59 \\ 
% \midrule 

% \multicolumn{7}{l}{\textit{Reference data}} \\ 
% Real shapes & 4{,}498 & 58 & 80 & 3 & 4{,}357 & 96.87 \\
% \bottomrule 
% \end{tabular} 
% \caption{Sequential verifier outcomes for real and generated shapes. Stage-specific flagging counts are conditional on passing all preceding filters. The augmented diffusion labels indicate the source of the verifier-accepted synthetic data used for augmentation: global PCA (Diffusion-GPCA), distributed PCA (Diffusion-DPCA), or diffusion self-augmentation (Diffusion-SelfAug).}
% \label{tab:filtering_results} 

% \end{table*}

\begin{table}[t]
\centering
{\small
\begin{tabular}{lrrr}
\toprule
\textbf{Method}
& $\boldsymbol{N}$
& \textbf{Accepted}
& \textbf{Rate (\%)} \\
\midrule

Global PCA
& 100{,}000 & 13{,}821 & 13.82 \\
Distributed PCA
& 45{,}000 & 30{,}081 & 66.85 \\
VAE
& 40{,}000 & 24{,}684 & 61.71 \\
Diffusion
& 40{,}000 & 37{,}411 & 93.53 \\

\midrule

Diffusion-GPCA
& 40{,}000 & 28{,}186 & 70.47 \\
Diffusion-DPCA
& 40{,}000 & 34{,}527 & 86.32 \\
Diffusion-SelfAug
& 40{,}000 & 37{,}837 & 94.59 \\

\midrule
\multicolumn{4}{l}{\textit{Reference data}} \\
Real shapes
& 4{,}498 & 4{,}357 & 96.87 \\

\bottomrule
\end{tabular}
}
\caption{Overall verifier acceptance for real and generated shapes.
$N$ denotes the initial number of shapes; stage-wise rejection counts are reported in Table~\ref{tab:filtering_results_full}.}
\label{tab:filtering_results}
\end{table}

\subsection{Human Evaluation of Generated Shapes and the Verifier}

Six blinded raters evaluated 400 outputs balanced across four diffusion variants and verifier decisions (50 verifier-accepted and 50 verifier-rejected for each model). Verifier-accepted outputs received positive
validity ratings in 83.4\% of judgments, compared with 53.7\% for
rejected outputs; the gap was present for every model (Table~\ref{tab:human_eval}). Among accepted outputs, Diffusion-DPCA received 84.2\% valid ratings, compared with 72.7\% for Diffusion-GPCA and 88.9\% for self-augmented diffusion. Per-shape majority votes agree with the verifier on 68.2\% of 365 decidable shapes ($\kappa=0.35$); inter-rater agreement is moderate ($\alpha=0.46$). 
The 53.7\% positive-rating rate among verifier-rejected shapes suggests that the verifier is conservative relative to visual inspection and may flag deviations that are difficult to assess from static renderings. Together with the moderate inter-rater agreement, this supports its use as a gatekeeper to reduce the risk of introducing implausible shapes into augmentation.

\begin{table}[t]
\centering
\small
\setlength{\tabcolsep}{4pt}
\begin{tabular}{@{}lccc@{}}
\toprule
Model & Verifier-Accepted & Verifier-Rejected & Gap\\
\midrule
Diffusion & 87.9 [83.7, 91.1] & 57.2 [51.5, 62.7] & +30.7 \\
Diffusion-GPCA & 72.7 [67.4, 77.4] & 32.4 [27.4, 37.9] & +40.2 \\
Diffusion-DPCA & 84.2 [79.6, 87.9] & 59.3 [53.7, 64.7] & +24.8 \\
Diffusion-SelfAug & 88.9 [84.9, 92.0] & 65.7 [60.1, 70.8] & +23.3 \\
\bottomrule
\end{tabular}
\caption{Positive human-validity ratings (\%) for verifier-accepted and rejected
outputs, with descriptive 95\% Wilson intervals. Full statistics appear in Table~\ref{tab:human_eval_full}.}
\label{tab:human_eval}
\end{table}

\subsection{Distributional Results}

\begin{figure}[t]
    \centering

    \begin{subfigure}{.39\textwidth}
        \centering
        \includegraphics[width=\linewidth]{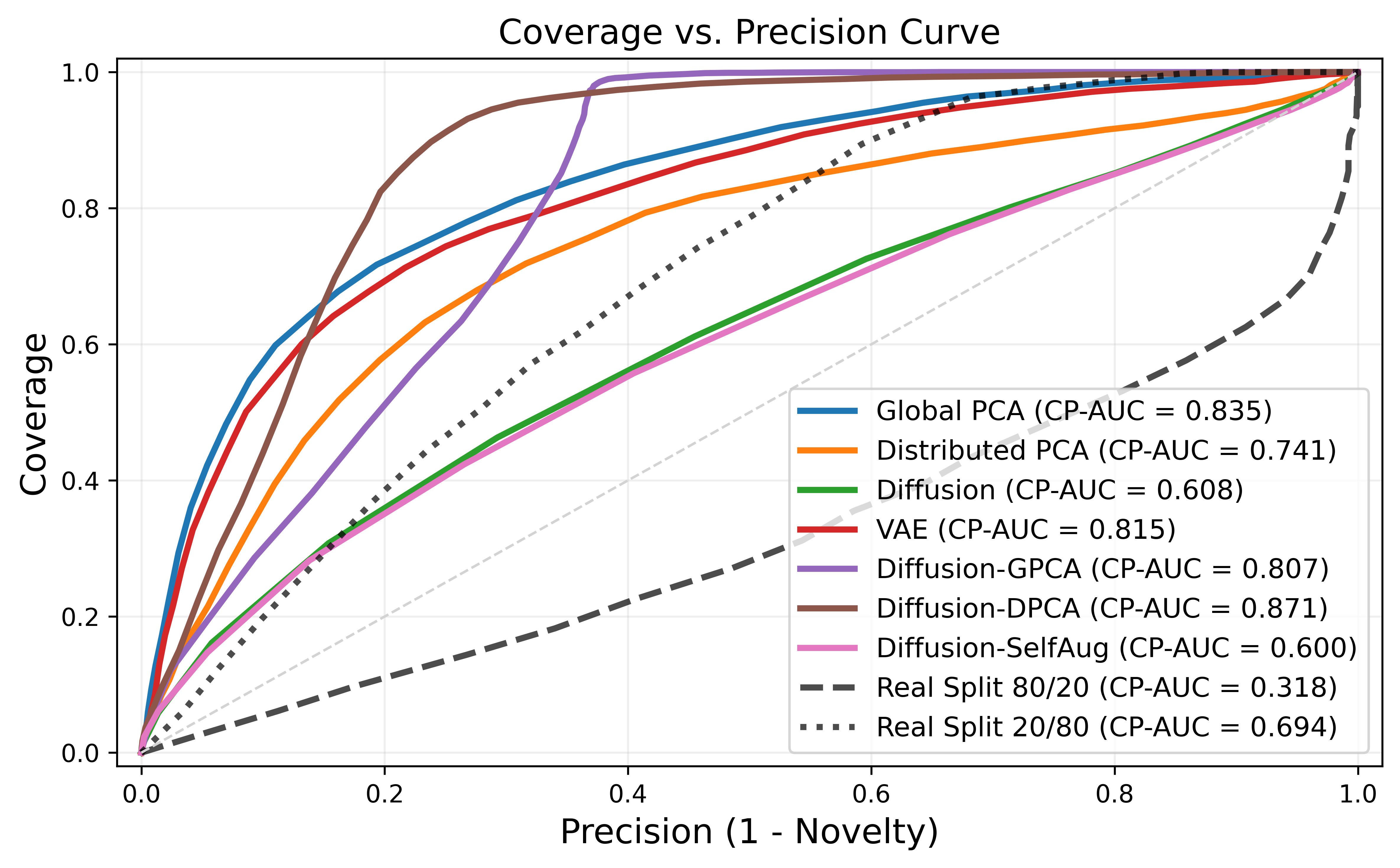}
        \subcaption{}
        \label{fig:cov_prec_ref}
    \end{subfigure}
    \par
    \begin{subfigure}{.22\textwidth}
        \centering
        \includegraphics[width=\linewidth]{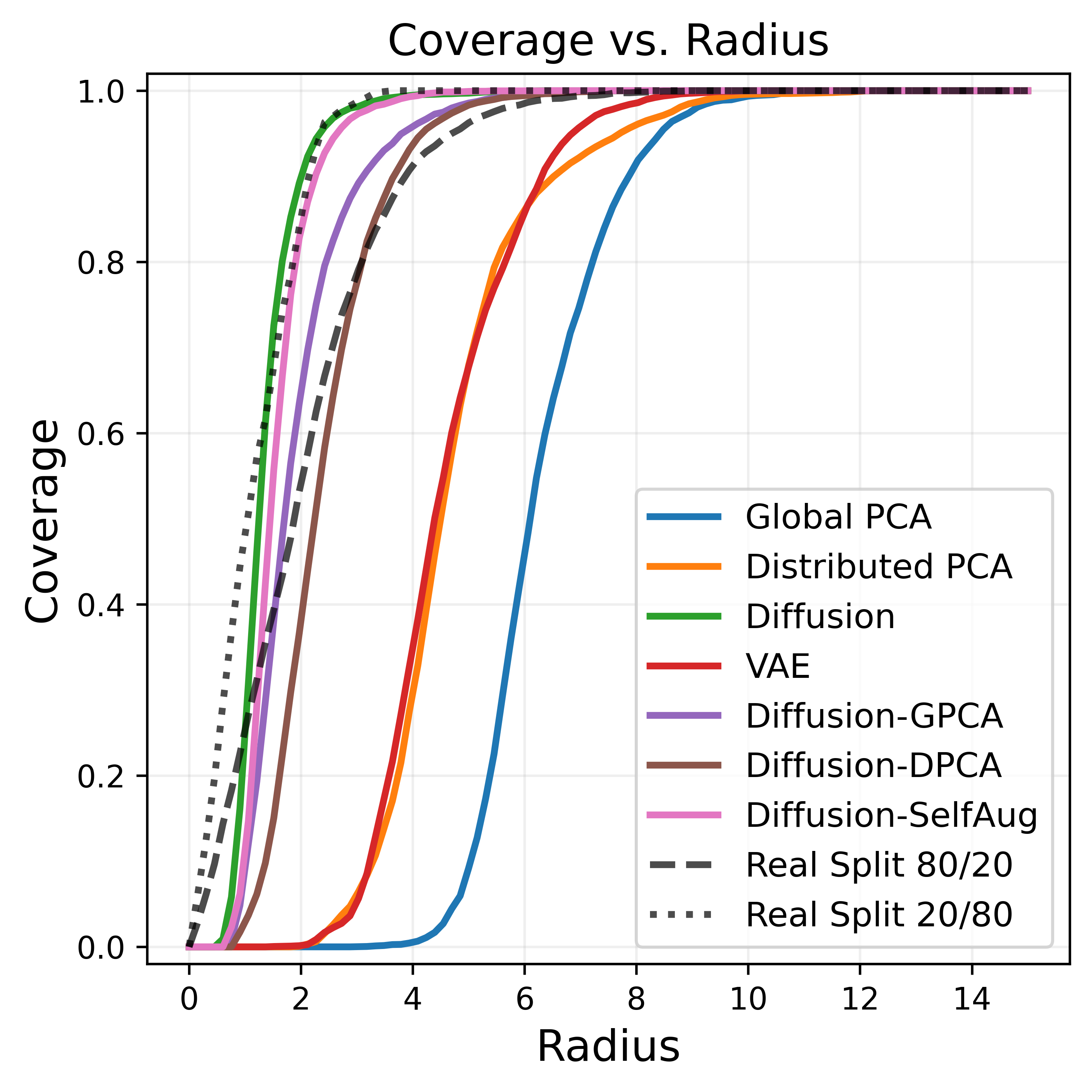}
        \subcaption{}
        \label{fig:cov_radi_ref}
    \end{subfigure}
    \hfill    
    \begin{subfigure}{.22\textwidth}
        \centering
        \includegraphics[width=\linewidth]{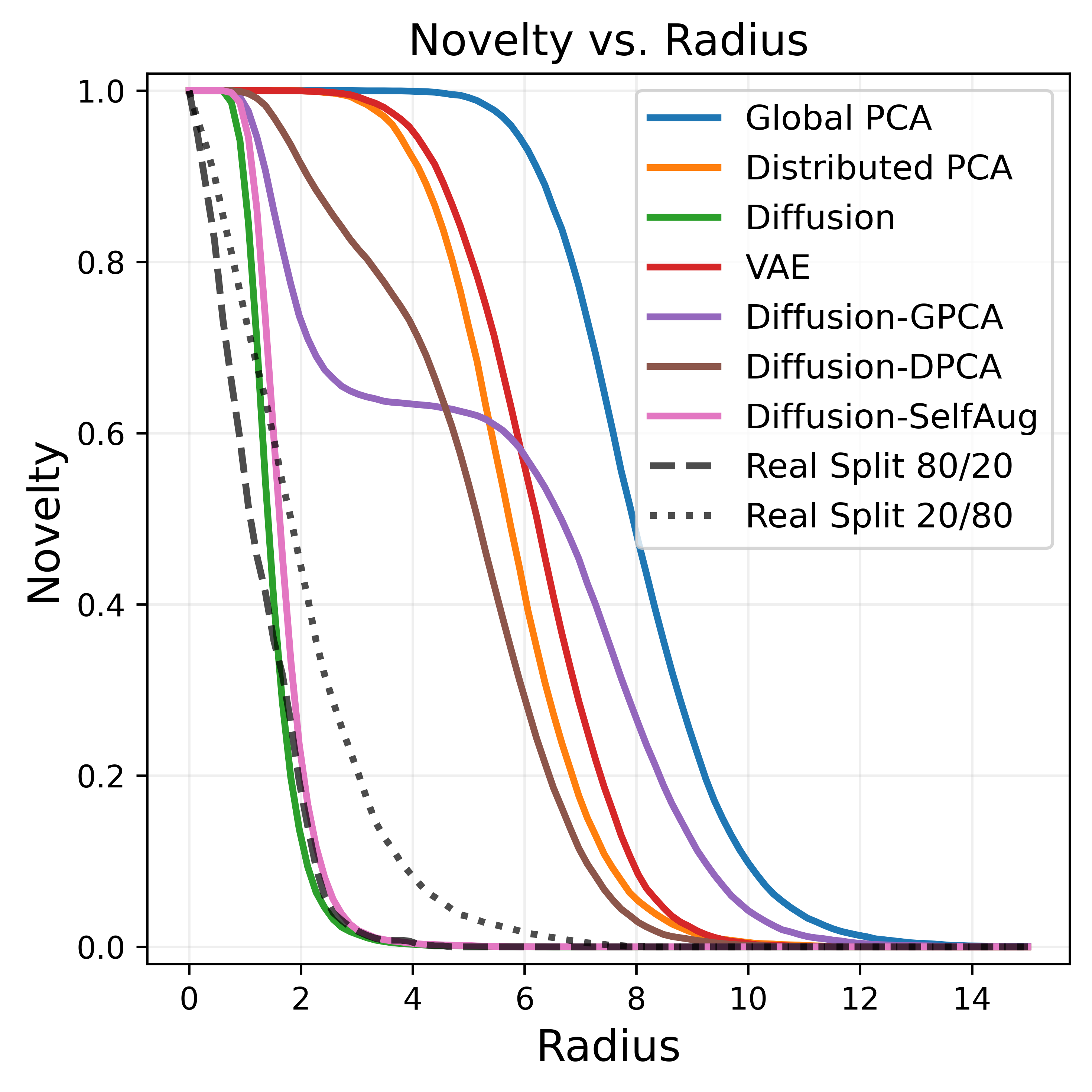}
        \subcaption{}
        \label{fig:nov_radi_ref}
    \end{subfigure}
    
    \caption{Distributional evaluation of 10,000 verifier-accepted shapes per method in the global PCA space. (a) Coverage--precision curves. (b) Coverage and (c) novelty versus radius. Dashed and dotted black curves denote real-split references.}
    \label{fig:distributional_evaluation_with_ref}
\end{figure}

We next examine whether high verifier acceptance is accompanied by meaningful distributional expansion. Fig.~\ref{fig:distributional_evaluation_with_ref} shows the coverage--precision, coverage--radius, and novelty--radius curves defined in Eqs.~\eqref{eq:Coverage}--\eqref{eq:Novelty}, using 10,000 randomly
selected verifier-accepted shapes per method. Table~\ref{tab:compact_summary} reports the corresponding quantile radii.

As shown in Fig.~\ref{fig:cov_radi_ref} and Fig.~\ref{fig:nov_radi_ref}, the real-only Diffusion model has the leftmost coverage curve, reaching each coverage level at the smallest coverage radii ($r_{50}^{\mathrm{cov}}=1.250$ and $r_{90}^{\mathrm{cov}}=2.009$). However, its novelty curve decays earliest and rapidly as radius increases, indicating close representation of the observed population but limited exploration
beyond it. Diffusion-SelfAug behaves similarly, which suggests that augmenting a diffusion model with its own accepted samples largely reinforces the learned distribution.

Global PCA exhibits the opposite pattern. Its coverage curve lies farthest to the right, but its accepted samples remain novel over the
largest radii. Distributed PCA and the VAE provide intermediate behavior: both cover the real population more closely than GPCA while retaining considerably greater novelty
than the real-only Diffusion model.

Diffusion retrained on verifier-accepted PCA samples exhibits broader accepted-sample support. Relative to the real-only model, both PCA-augmented models' novelty curves are shifted to the right. Diffusion-GPCA exhibits a plateau in its novelty curve in Fig.~\ref{fig:nov_radi_ref}, consistent with the large gap between $r_{50}^{\mathrm{nov}}=6.654$ and $r_{70}^{\mathrm{nov}}=2.198$; this suggests a heterogeneous mixture of
samples near the real data and samples much farther away. Diffusion-DPCA produces a more balanced shift: its coverage radii remain substantially smaller than those of the PCA proposal models, while its novelty radii are markedly larger than those of real-only diffusion. Consequently, the Diffusion-DPCA curve occupies the upper-left region of the coverage--precision plot in Fig.~\ref{fig:cov_prec_ref} over much of the precision range and achieves the highest CP-AUC of $0.871$, compared with $0.608$ for Diffusion and $0.807$ for Diffusion-GPCA. These results suggest that Diffusion-DPCA does not simply move accepted samples away from the observations: it retains substantially smaller coverage radii than the PCA proposal models while generating verifier-accepted samples farther from the observed set than real-only diffusion. Using random partitions of the real population as a calibration, Diffusion-DPCA's CP-AUC of 0.871 closely matches the 0.872 obtained for a 5\% target/95\% pseudo-generated split (Table~\ref{tab:real_split_ref}), indicating substantial expansion into regions of shape space not represented by the observed sample.

\begin{table}[t]
\centering
\small
\setlength{\tabcolsep}{1mm}
\begin{tabular}{lcccccc}
\toprule
% & \multicolumn{4}{c}{AUC} 
& \multicolumn{3}{c}{Coverage radii} 
% & \multicolumn{3}{c}{Precision radii} 
& \multicolumn{3}{c}{Novelty radii} \\
% \cmidrule(lr){2-4} \cmidrule(lr){5-7} \cmidrule(lr){8-10} \cmidrule(lr){12-14}
Method 
% & Cov & Prec & Nov & PR 
& $r_{50}^{\mathrm{cov}}$ & $r_{70}^{\mathrm{cov}}$ & $r_{90}^{\mathrm{cov}}$ 
% & $r_{50}$ & $r_{70}$ & $r_{90}$ 
& $r_{50}^{\mathrm{nov}}$ & $r_{70}^{\mathrm{nov}}$ & $r_{90}^{\mathrm{nov}}$ \\
\midrule
Global PCA 
& 6.101 & 6.753 & 7.865 
% & 7.938 & 8.733 & 9.981 
& \textbf{7.938} & \textbf{7.242} & \textbf{6.290} \\
 
Distributed PCA 
& 4.499 & 5.080 & 6.524 
% & 5.725 & 6.401 & 7.502 
& 5.725 & 5.091 & 4.174 \\

VAE
& 4.392 & 5.097 & 6.309 
% & 6.221 & 6.919 & 7.925 
& 6.221 & 5.506 & 4.491 \\
 
Diffusion 
& \textbf{1.250} & \textbf{1.481} & \textbf{2.009} 
% & \textbf{1.412} & \textbf{1.650} & \textbf{2.099} 
& 1.412 & 1.219 & 0.976 \\
 
Diffusion-GPCA
& 1.705 & 2.129 & 3.114 
% & 6.654 & 7.807 & 9.219 
& 6.654 & 2.198 & 1.386 \\

Diffusion-DPCA
& 2.248 & 2.734 & 3.661 
% & 5.160 & 5.961 & 7.096 
& 5.160 & 4.175 & 2.123 \\

Diffusion-SelfAug
& 1.450 & 1.720 & 2.259 
% & 1.621 & 1.873 & 2.345 
& 1.621 & 1.401 & 1.143\\

\bottomrule
\end{tabular}
\caption{Coverage and novelty quantile radii in global PCA space; bold marks the column optimum.}

\label{tab:compact_summary}
\end{table}

Within the controlled 10,000-sample comparison, Diffusion-DPCA provides the strongest coverage--novelty balance among the augmented diffusion models; the real-split curves provide the corresponding finite-sample calibration shown in Table~\ref{tab:real_split_ref}.

% \begin{table} 
%     \centering
%     \small
%     \setlength{\tabcolsep}{4pt}
%     \begin{tabular}{lccc}
%     \toprule
%     \textbf{Split (\%)} & \textbf{Target} & \textbf{Pseudo-generated} & \textbf{CP-AUC} \\
%     \midrule
%     95/5  & 4,273 &   225 & 0.122 \\
%     90/10 & 4,048 &   450 & 0.214 \\
%     80/20 & 3,598 &   900 & 0.318 \\
%     70/30 & 3,148 & 1,350 & 0.394 \\
%     60/40 & 2,698 & 1,800 & 0.461 \\
%     50/50 & 2,249 & 2,249 & 0.506 \\
%     40/60 & 1,799 & 2,699 & 0.565 \\
%     30/70 & 1,349 & 3,149 & 0.623 \\
%     20/80 &   899 & 3,599 & 0.694 \\
%     10/90 &   449 & 4,049 & 0.797 \\
%     5/95  &   224 & 4,274 & 0.872 \\
%     \bottomrule
%     \end{tabular}
%     \caption{Reference benchmark CP-AUC values obtained by randomly splitting the real shape population into two subsets (target and pseudo-generated).}
%     \label{tab:real_split_ref}
% \end{table}

\subsection{Acceptance-Adjusted Performance}
We summarize the acceptance--exploration trade-off using EAUC (Eq.~\eqref{eq:eauc}). In Table~\ref{tab:eauc}, Diffusion and Diffusion-SelfAug exceed 93\% acceptance but remain concentrated near the observations, while Diffusion-GPCA increases CP-AUC at lower acceptance. Diffusion-DPCA combines 86.32\% acceptance with the highest CP-AUC (0.871) and EAUC (0.752), a 32\% increase over real-only and self-augmented Diffusion. Together with the distance curves, these results identify Diffusion-DPCA as providing the strongest acceptance--coverage--departure trade-off in our comparison.

\begin{table}[t]
    \centering
    \small
    \begin{tabular}{lcc|c}
    \toprule
    \textbf{Method} & \textbf{Acceptance (\%)} & \textbf{CP-AUC} & \textbf{EAUC} \\
    \midrule
    Global PCA & 13.82 & 0.835 & 0.115 \\
    Distributed PCA & 66.85 & 0.741 & 0.495 \\
    VAE & 61.71 & 0.815  & 0.503 \\
    Diffusion & 93.53 & 0.608 & 0.569 \\
    Diffusion-GPCA & 70.47 & 0.807 & 0.569 \\
    Diffusion-DPCA & 86.32 & \textbf{0.871} & \textbf{0.752} \\
    Diffusion-SelfAug & \textbf{94.59} & 0.600 & 0.568 \\
    
    \bottomrule
    \end{tabular}
    \caption{Verifier acceptance over full candidate pools and CP-AUC over 10,000 verifier-accepted shapes; EAUC is their product. Bold indicates the highest value in each column.}
    \label{tab:eauc}
\end{table}

\begin{figure}[t]
    \centering
    \includegraphics[width=.9\linewidth]{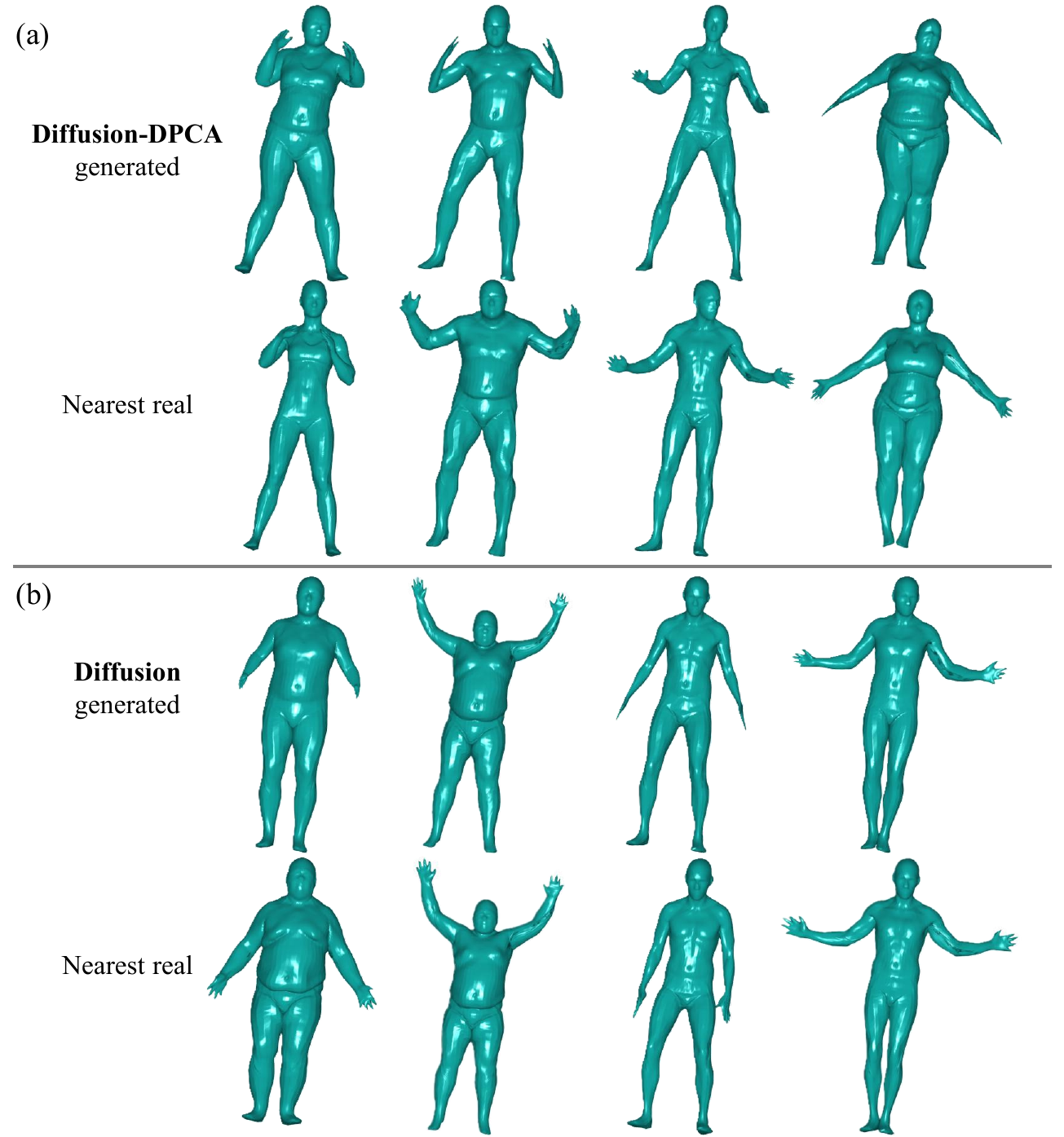}
    \caption{Examples of highly novel, accepted shapes shown above their nearest real neighbors: (a) Diffusion-DPCA. (b) Real-only Diffusion. Examples are randomly selected from the top 1\% most novel samples by nearest-real distance.}
    \label{fig:novel_shape}
\end{figure}

\paragraph{Qualitative Examples.}
The acceptance-adjusted results identify Diffusion-DPCA as providing a balance between verifier acceptance and distributional expansion. Fig.~\ref{fig:novel_shape} provides a qualitative view of this result. For each method, we rank 10,000 verifier-accepted samples by nearest-real distance and randomly select four from the top 1\% most novel samples. Compared with their real neighbors, Diffusion-DPCA examples exhibit more pronounced variation in body morphology and pose while passing the same verifier, whereas the real-only Diffusion examples generally remain closer to observed shapes. Additional verifier-accepted novel shapes are shown in Fig.~\ref{fig:novel_shapes_add} in the Appendix.

\section{Conclusion and Future Work}
We introduced a verifier-guided proposal--verification--augmentation framework for 3D human-shape generative models under limited data. Global and distributed PCA provide low-cost candidate exploration, elastic correspondence enables scalable screening using skeletal and body-part measurements, and diffusion learns a refined distribution from the real and verifier-accepted synthetic data. We further introduced an acceptance-aware evaluation
that jointly reports full-pool verifier acceptance and the coverage--novelty behavior of accepted samples. On DFAUST, DPCA-augmented Diffusion attains the highest CP-AUC ($0.871$) and EAUC ($0.752$), delivering the strongest acceptance--coverage--departure balance in our comparison. Blinded human ratings further support the verifier as a practical, scalable complement to human assessment: accepted outputs were judged valid substantially more often than rejected outputs, while moderate inter-rater agreement highlights the ambiguity of human visual assessment.

The focused 10-subject, two-motion setting demonstrates the framework in the limited, concentrated populations common to scientific shape-analysis applications. Future work will evaluate robustness on the full DFAUST collection and larger datasets, and extend the verifier with other plausibility checks. These extensions may benefit downstream applications such as body reconstruction, pose estimation, ergonomic modeling, and underrepresented body morphology learning. More broadly, the proposed strategy may apply to other structured-data domains in which large candidate pools can be generated inexpensively and a reference-calibrated screening rule is available to guide synthetic augmentation.

\bibliography{arxiv2027}

\clearpage
\onecolumn
\appendix
\renewcommand{\thetable}{A\arabic{table}}
\renewcommand{\thefigure}{A\arabic{figure}}
\setcounter{table}{0}
\setcounter{figure}{0}

\section{Appendix}
\begin{figure}[htbp]
    \centering
    \includegraphics[width=.85\linewidth]{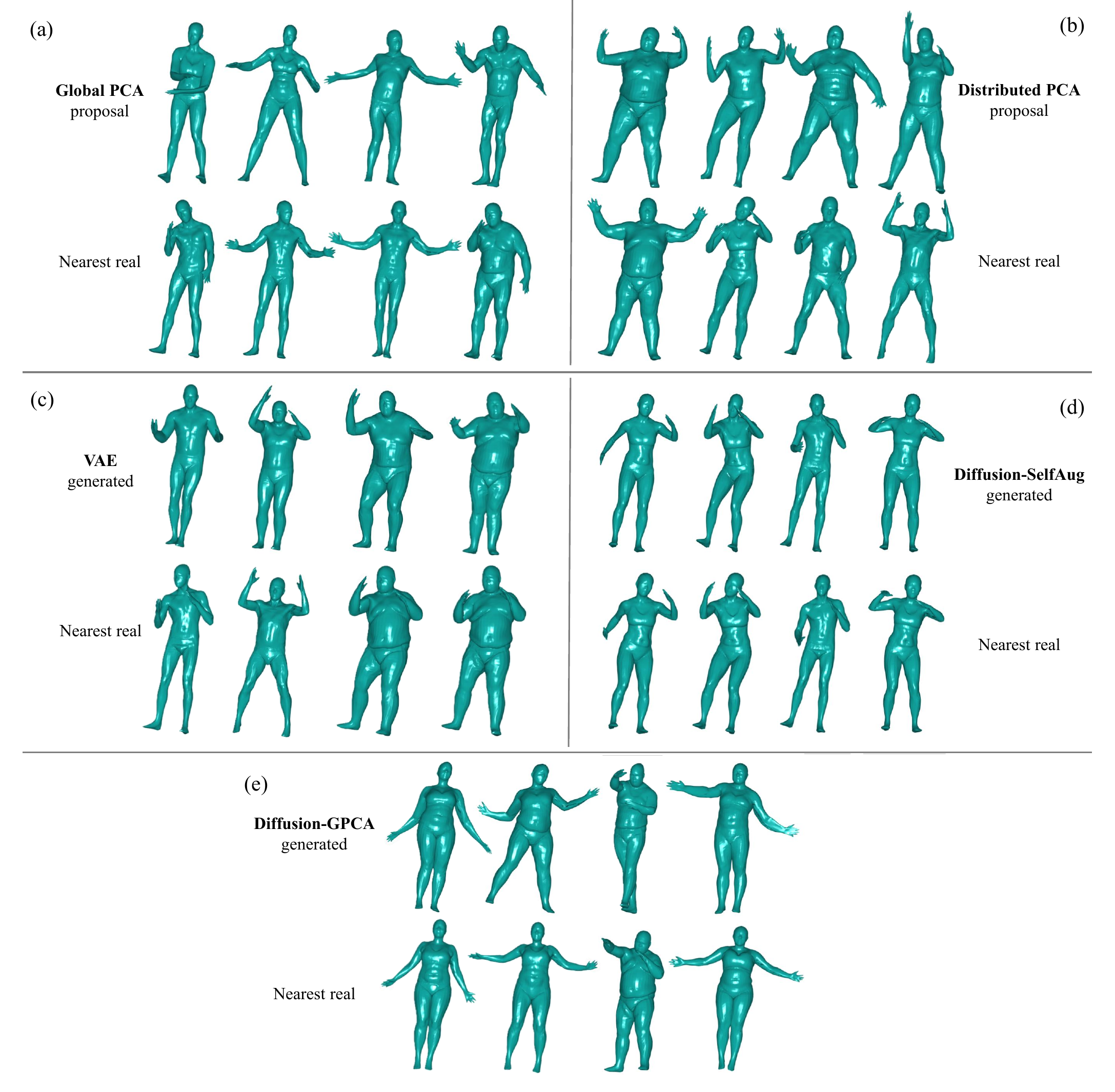}
    \caption{Additional highly novel verifier-accepted outputs shown above their nearest real neighbors. Four outputs per method are randomly selected from the top 1\% by nearest-real distance: (a) GPCA; (b) DPCA; (c) VAE; (d) Diffusion-SelfAug; and (e) Diffusion-GPCA.}
    \label{fig:novel_shapes_add}
\end{figure}

\vspace{.5cm}
\paragraph{Additional Novel Accepted Shapes.}
Fig.~\ref{fig:novel_shapes_add} complements Fig.~\ref{fig:novel_shape} with highly novel verifier-accepted outputs from the remaining methods, each shown above its nearest real neighbor.

\newpage

\paragraph{Stage-Wise Verifier Outcomes.}
Table~\ref{tab:filtering_results_full} expands Table~\ref{tab:filtering_results} by reporting, for each source, the number of shapes newly rejected at the length, volume, and arm-area stages. Counts at each stage are conditional on passing all preceding stages.

\begin{table}[htbp]
\centering 

\small
\begin{tabular}{lrrrrrr} 

\toprule 

& \textbf{Initial number} 

& \multicolumn{3}{c}{\textbf{Flagged by filtering metric}} 

& \textbf{Verifier} 

& \textbf{Acceptance} \\
\cmidrule(lr){3-5} \textbf{Source} & \textbf{of shapes} & \textbf{Length} & \textbf{Volume} & \textbf{Arm-area} & \textbf{accepted} & \textbf{rate (\%)} \\
\midrule

\multicolumn{7}{l}{\textit{Base generators}} \\ 

Global PCA & 100{,}000 & 80{,}737 & 5{,}204 & 238 & 13{,}821 & 13.82 \\ Distributed PCA & 45{,}000 & 11{,}648 & 2{,}664 & 607 & 30{,}081 & 66.85 \\ 

VAE & 40{,}000 & 13{,}304 & 1{,}768 & 244 & 24{,}684 & 61.71 \\
Diffusion & 40{,}000 & 1{,}763 & 697 & 129 & 37{,}411 & 93.53 \\
\midrule

\multicolumn{7}{l}{\textit{Augmented diffusion generators}} \\
Diffusion-GPCA & 40{,}000 & 9{,}250 & 2{,}285 & 279 & 28{,}186 & 70.47 \\
Diffusion-DPCA & 40{,}000 & 3{,}875 & 1{,}243 & 355 & 34{,}527 & 86.32 \\ Diffusion-SelfAug & 40{,}000 & 1{,}044 & 1{,}041 & 78 & 37{,}837 & 94.59 \\ 

\midrule

\multicolumn{7}{l}{\textit{Reference data}} \\ 

Real shapes & 4{,}498 & 58 & 80 & 3 & 4{,}357 & 96.87 \\
\bottomrule 

\end{tabular} 

\caption{Detailed sequential verifier outcomes for real and generated shapes. Length-, volume-, and arm-area-specific counts report shapes
newly rejected after passing all preceding filtering stages.}
\label{tab:filtering_results_full} 
\end{table}

\vspace{.7cm}
\paragraph{Real-Split CP-AUC Calibration.}
Table~\ref{tab:real_split_ref} reports CP-AUC when the 4,498 real surfaces are randomly partitioned into a target subset and a pseudo-generated subset at varying proportions.

\begin{table}[htbp]
    \centering
    \small
    \setlength{\tabcolsep}{4pt}
    \begin{tabular}{lccc}
    \toprule
    \textbf{Split (\%)} & \textbf{Target} & \textbf{Pseudo-generated} & \textbf{CP-AUC} \\
    \midrule
    95/5  & 4,273 &   225 & 0.122 \\
    90/10 & 4,048 &   450 & 0.214 \\
    80/20 & 3,598 &   900 & 0.318 \\
    70/30 & 3,148 & 1,350 & 0.394 \\
    60/40 & 2,698 & 1,800 & 0.461 \\
    50/50 & 2,249 & 2,249 & 0.506 \\
    40/60 & 1,799 & 2,699 & 0.565 \\
    30/70 & 1,349 & 3,149 & 0.623 \\
    20/80 &   899 & 3,599 & 0.694 \\
    10/90 &   449 & 4,049 & 0.797 \\
    5/95  &   224 & 4,274 & 0.872 \\
    \bottomrule
    \end{tabular}
    \caption{Complete CP-AUC sensitivity analysis under target/pseudo-generated partitions of the 4,498 real surfaces. Split percentages are reported as target/pseudo-generated.}
    \label{tab:real_split_ref}
\end{table}

\newpage
\paragraph{Full Human-Evaluation Statistics.}
Table~\ref{tab:human_eval_full} reports the full human-evaluation statistics for each generation setting and verifier decision.

\begin{table}[htbp]
\centering
\small
\setlength{\tabcolsep}{4.5pt}
\begin{tabular}{l ccc ccc c}
\toprule
& \multicolumn{3}{c}{Verifier-accepted} & \multicolumn{3}{c}{Verifier-rejected} & \\
\cmidrule(lr){2-4}\cmidrule(lr){5-7}
Setting & Valid (\%) & Wilson & Bootstrap & Valid (\%) & Wilson & Bootstrap & Gap \\
\midrule
Real-only diffusion & 87.9 {\scriptsize(262/298)} & [83.7, 91.1] & [79.0, 94.9] & 57.2 {\scriptsize(171/299)} & [51.5, 62.7] & [53.7, 60.3] & +30.7 \\
Diffusion-GPCA & 72.7 {\scriptsize(218/300)} & [67.4, 77.4] & [62.7, 80.7] & 32.4 {\scriptsize(97/299)}  & [27.4, 37.9] & [27.4, 37.2] & +40.2 \\
Diffusion-DPCA & 84.2 {\scriptsize(250/297)} & [79.6, 87.9] & [78.5, 89.6] & 59.3 {\scriptsize(178/300)} & [53.7, 64.7] & [53.3, 65.0] & +24.8 \\
Diffusion-SelfAug & 88.9 {\scriptsize(265/298)} & [84.9, 92.0] & [84.2, 94.0] & 65.7 {\scriptsize(197/300)} & [60.1, 70.8] & [58.0, 71.0] & +23.3 \\
\midrule
Overall              & 83.4 {\scriptsize(995/1193)} & [81.2, 85.4] & [76.4, 89.5] & 53.7 {\scriptsize(643/1198)} & [50.8, 56.5] & [48.5, 57.8] & +29.7 \\
\bottomrule
\end{tabular}
\caption{
Full human-evaluation statistics: share of judgments rating shapes valid human bodies, by generation setting and verifier decision, with descriptive 95\% Wilson
intervals and 95\% percentile intervals from a rater-cluster bootstrap (raters resampled
with replacement, $B{=}10{,}000$). Gap = accepted minus rejected in percentage points.}
\label{tab:human_eval_full}
\end{table}

\end{document}